\documentclass[runningheads]{llncs}
\usepackage{graphicx}
\usepackage{amsmath,amssymb} % define this before the line numbering.
\usepackage{color}
\usepackage[width=122mm,left=12mm,paperwidth=146mm,height=193mm,top=12mm,paperheight=217mm]{geometry}
\usepackage{authblk}
\usepackage[noend]{algpseudocode}
\usepackage{algorithmicx,algorithm}
\usepackage{enumitem}
\usepackage{bbm}
\begin{document}
% \linenumbers
%\pagestyle{headings}
%\mainmatter

\title{Selective Zero-Shot Classification with Augmented Attributes}
% Replace with your title

\titlerunning{Selective Zero-Shot Classification}
% Replace with a meaningful short version of your title

\authorrunning{J. Song et al.}
% Replace with shorter version of the author list. If there are more authors than fits a line, please use A. Author et al.

\author{Jie Song \inst{1} \and Chengchao Shen \inst{1} \and Jie Lei \inst{1} \and An-Xiang Zeng \inst{2} \and Kairi Ou \inst{2} \and Dacheng Tao \inst{3} \and Mingli Song\inst{1}}

%Please write out author names in full in the paper, i.e. full given and family names.
%If any authors have names that can be parsed into FirstName LastName in multiple ways, please include the correct parsing, in a comment to the volume editors:
%\index{Lastnames, Firstnames}
%(Do not uncomment it, because you may introduce extra index items if you do that, we will use scripts for introducing index entries...)

\institute{College of Computer Science and Technology, Zhejiang University, Hangzhou, China\\
    \email{ \{sjie,chengchaoshen,ljaylei,brooksong\}@zju.edu.cn}\\
	\and Alibaba Group, Hangzhou, China\\
    \email{ \{renzhong,suzhe.okr\}@taobao.com}\\
    \and UBTECH Sydney AI Centre, SIT, FEIT, University of Sydney, Australia\\
    \email{ dacheng.tao@sydney.edu.au}}
\maketitle

\begin{abstract}
In this paper, we introduce a selective zero-shot classification problem: how can the classifier avoid making dubious predictions? Existing attribute-based zero-shot classification methods are shown to work poorly in the selective classification scenario. We argue the under-complete human defined attribute vocabulary accounts for the poor performance. We propose a selective zero-shot classifier based on both the human defined and the automatically discovered residual attributes. The proposed classifier is constructed by firstly learning the defined and the residual attributes jointly. Then the predictions are conducted within the subspace of the defined attributes. Finally, the prediction confidence is measured by both the defined and the residual attributes. Experiments conducted on several benchmarks demonstrate that our classifier produces a superior performance to other methods under the risk-coverage trade-off metric.

\end{abstract}

\section{Introduction}

Zero-Shot Classification (ZSC) addresses the problem of recognizing images from novel categories, \textit{i.e.}, those categories which are not seen during the training phase. It has attracted much attention~\cite{lampert2009learning,akata2013label,romera2015embarrassingly,akata2015evaluation,Xian_2017_CVPR,Song_2018_CVPR} in the last decade due to its importance in real-world applications, where the data collection and annotation are both laboriously difficult. Existing ZSC methods usually assume that both the seen and the unseen categories share a common semantic space (\textit{e.g.}, attributes~\cite{lampert2009learning,akata2013label}) where both the images and the class names can be projected. Under this assumption, the recognition of images from unseen categories can be achieved by the nearest neighbor search in the shared semantic space.

Although there is a large literature on ZSC, the prediction of existing zero-shot classifiers remains quite unreliable compared to that of the fully supervised classifiers. This limits their deployment in real-world applications, especially where mistakes may cause severe risks. For example, in autonomous driving, a wrong decision can result in traffic accidents. In clinical trials, a misdiagnosis may make the patient suffer from great pain and loss.

To reduce the risk of misclassifications, selective classification improves classification accuracy by rejecting examples that fall below a confidence threshold~\cite{Chow1957,El-Yaniv:2010:FNS:1756006.1859904}. Motivated by this, in this paper we introduce a Selective Zero-Shot Classification (Selective ZSC) problem: the zero-shot classifier can abstain from predicting when it is uncertainty about its predictions. It requires that the classifier not only makes accurate predictions given images from unseen categories but also be self-aware. In other words, the classifier should be able to know when it is confident (or uncertain) about their predictions. The confidence is typically quantified by a confidence score function. Equipped with this ability, the classifier can leave the classification of images when it is uncertain about its predictions to the external domain expert (\textit{e.g.,} drivers in autonomous driving, or doctors in clinical trials).

Selective classification is an old topic in machine learning field. However, we highlight its importance in the context of ZSC in threefold. Firstly, the predictions of zero-shot classifiers are not so accurate compared with those of fully supervised classifiers, which poses large difficulty in Selective ZSC. Secondly, it is shown in our experiments (in Section~\ref{ex:benchmark}) that most existing zero-shot classifiers exhibit poor self-awareness. This results in their inferior performance in the settings of Selective ZSC. Lastly, albeit its great importance in real-world applications, selective classification remains under-studied in the field of ZSC.

Typically, existing ZSC methods rely on human defined attributes for novel class recognition. Attributes are a type of mid-level semantic properties of visual objects that can be shared across different object categories. Manually defined attributes are often those nameable properties such as color, shape, and texture. However, the discriminative properties for the classification task are often not exhaustively defined and sometimes hard to be described in a few words or some semantic concepts. Thus, the under-complete defined attribute vocabulary results in inferior performance of attribute-based ZSC methods. We call the residual discriminative but not defined properties \textit{residual attributes}. To make safer predictions for zero-shot classification, we argue both the defined and the residual attributes should be exploited. These two types of attributes together are named \textit{augmented attributes} in this paper.

We propose a much safer selective classifier for zero-shot recognition based on augmented attributes. The proposed classifier is constructed by firstly learning the augmented attributes. Motivated by~\cite{Peng2016,Jiang_2017_ICCV}, we formulate the attribute learning task as a dictionary learning problem. After the learning of the augmented attributes, the defined attributes can be directly utilized to accomplish traditional zero-shot recognitions. The confidence function thus can be defined within the subspace of defined attributes.  The residual attributes, however, can not be directly exploited for classification because there are no associations between the residual attributes and the unseen categories. Instead of conducting direct predictions, we leverage the residual attributes to improve the self-awareness of the classifier constructed on defined attributes. Specifically, we define another confidence function based on the consistency between the defined and the residual attributes. Combining the confidence obtained on the augmented attributes and confidence produced within the defined attributes, the proposed selective classifier significantly outperforms other methods in extensive experiments.

To sum up, we made the following contributions: (1) we introduce the selective zero-shot classification problem, which is important yet under-studied; (2) we propose a selective zero-shot classifier, which leverages both the manually defined and the automatically discovered residual attributes for safer predictions; (3) we propose a solution to the learning of residual discriminative properties in addition to the manually defined attributes; (4) experiments demonstrate our method significantly outperforms existing state-of-the-art methods.

\section{Related Work}

\subsection{Zero-Shot Learning}
Typically, existing ZSC methods consist of two steps. The first step is an embedding process, which maps both the image representations and the class names to a shared embedding space. This step can also be viewed as a kind of multi-modality matching problem~\cite{Kan_2016_CVPR,Transductive_Multi_view}. The second step is a recognition process, which is usually accomplished by some form of nearest neighbor searches in the shared space learned from the first step. Existing ZSC approaches mainly differ in the choices for the embedding model and the recognition model. For example, DAP~\cite{lampert2009learning} adopts probabilistic attribute classifiers for embedding and Bayes classifier for recognition. Devise~\cite{frome2013devise}, Attribute Label Embedding (ALE)~\cite{Akata_2016}, Simple ZSC~\cite{romera2015embarrassingly} and Structured Joint Embedding (SJE)~\cite{akata2015evaluation} adopt linear projection and inner product for embedding and recognition, respectively. However, they exploit different objective functions for optimization. Embedding Model (LatEm)~\cite{Xian_2016_CVPR} and Cross Model Transfer (CMT)~\cite{socher2013zero} employ nonlinear projection for embedding to overcome the limitations of linear models. Different from above methods, Semantic Similarity Embedding (SSE)~\cite{zhang2015zero}, Convex Combination of Semantic Embeddings (CONSE)~\cite{norouzi2014zeroshot} and Synthesized Classifiers (SYNC)~\cite{changpinyo2016synthesized} build the shared embedding space by expressing images and semantic class embeddings as a mixture of seen class proportions. For a more comprehensive review about ZSC, please refer to~\cite{Recent_Advances,Xian_2017_CVPR}.

\subsection{Defined Attributes and Latent Attributes}
Attributes are usually defined as the explainable properties such as color, shape, and parts. With manually defined attributes as a shared semantic vocabulary, novel classes can be easily defined such that zero-shot recognition can be accomplished via the association between the defined attributes and the categories. However, manually finding a discriminative and meaningful set of attributes can sometimes be difficult. The method for learning discriminative latent attributes has been exploited~\cite{Berg2010,Sharmanska2012,Rastegari2012,Category-LevelAttributes2013,Peng2016}. Tamara \textit{et al.}~\cite{Berg2010} propose to automatically identify attributes vocabulary from text descriptions of images sampled from the Internet. Viktoriia \textit{et al.}~\cite{Sharmanska2012} propose to augment defined attributes with latent attributes to facilitate few-shot learning. Mohammad \textit{et al.}~\cite{Rastegari2012} propose to discover attributes by trading off between predictability and discrimination. Felix \textit{et al.}~\cite{Category-LevelAttributes2013} propose to design attributes without concise semantic terms for visual recognition by incorporating both the category-separability and the learnability into the learning criteria. Peixi \textit{et al.}~\cite{Peng2016} propose a dictionary learning model to decompose the dictionary space into three parts corresponding to defined, latent discriminative and latent background attributes. Different from these works, in this paper we augment the manually defined attributes with residual attributes to improve the self-awareness of zero-shot classifier.

\subsection{Selective Classification}
Safety issues have attracted much attention in the AI research community in the last several years. For example, Szegedy \textit{et al.}~\cite{Szegedy2014} find that deep neural networks are easily fooled by adversarial examples. Following their work, many methods are proposed to construct more robust classifiers.

To reduce the risk of misclassifications, selective classification~\cite{Chow1957,El-Yaniv:2010:FNS:1756006.1859904} improve classification accuracy by rejecting examples that fall below a confidence threshold. For different classifiers, the confidence scores can be defined in various ways. Most generative classification models are probabilistic, therefore they provide such confidence scores in nature. However, most discriminative models do not have direct access to the probability of their predictions~\cite{DBLP:journals/corr/abs-1709-09844}. Instead, related non-probabilistic scores are used as proxies, such as the margin in the SVM classifier and the softmax output or MC-Dropout~\cite{Gal2016Dropout} in deep neural networks. In this paper, we propose to exploit the residual attributes to compensate the limitations of defined attributes and make the classifier more self-aware.

\section{Problem Formulation of Selective Zero-Shot Classification}
We summarize some key notations used in this paper in Table 1 for reference.

\begin{table}[ht]
  \caption{Some key notations used in this paper. Some of them are also explained in the main text. }
  \label{table:notations}
  \centering
  \begin{tabular}{|l||l|}
  \hline
      \textbf{Notations}           & \textbf{Definition}\\
      \hline
      $\mathcal{Y}_s, \mathcal{Y}_u$       & Seen label set and unseen label set\\
      $N_s, N_u$    & Number of seen (unseen) images, $N_s$ $(N_u)\in\mathbb{N}^+$\\
      $K_o$    & Number of dimensions of the feature space, $K_o\in\mathbb{N}^+$\\
      $\textbf{x}_i$    & An instance in the feature space, $\textbf{x}_i\in \mathbb{R}^{K_o}$\\
      $\textbf{X}_s, \textbf{X}_u$    & Seen/Unseen image representations, $\textbf{X}_s\in \mathbb{R}^{K_o \times N_s}, \textbf{X}_u\in \mathbb{R}^{K_o \times N_u}$ \\
      $\textbf{y}_s$    & Label annotations for the training data $\textbf{X}_s$, $\textbf{y}_s\in \mathbb{R}^{N_s}$\\
      \hline
      \hline
      $K_d, K_r$    & Number of dimensions of the defined and the residual attribute space\\
      $\textbf{D}_s$    & Defined attribute annotations $\textbf{D}_s\in\mathbb{R}^{K_d\times N_s}$ for the training data $\textbf{X}_s$\\
      $\textbf{D}_o$    & Defined attribute annotations $\textbf{D}_o\in\mathbb{R}^{K_d\times|\mathcal{Y}_s|}$ for the seen classes \\
      $\textbf{R}_o$    & Residual attribute representations $\textbf{R}_o\in\mathbb{R}^{K_r\times|\mathcal{Y}_s|}$ for the seen classes\\
      $[\textbf{d}_i; \textbf{r}_i]$ & Augmented attribute representation of $\textbf{x}_i$. $\textbf{d}_i$ is the defined attributes, \\
      & and $\textbf{r}_i$ is the residual attributes\\
      $[\textbf{d}^j; \textbf{r}^j]$  &Augmented attribute representation of class $j$\\
      $\textbf{s}_d, \textbf{s}_r$    & Similarity vectors from the defined/residual attributes, $\textbf{s}_d, \textbf{s}_r\in\mathbb{R}^{|\mathcal{Y}_s|}$\\
      \hline
  \end{tabular}\\
\end{table}

Let $\mathcal{X}$ be the feature space (e.g., raw image data or feature vectors) and $\mathcal{Y}$ be a finite label set. Let $P_{\mathcal{X},\mathcal{Y}}$ be a distribution over $\mathcal{X}\times\mathcal{Y}$. In a standard multi-class zero-shot classification problem, given training data $\textbf{X}_s=[\textbf{x}_1, \textbf{x}_2, ..., \textbf{x}_{N_s}]$ and corresponding defined attribute annotations $\textbf{D}_s=[\textbf{d}_1, \textbf{d}_2, ..., \textbf{d}_{N_s}]$ and label annotations $\textbf{y}_s=[y_1, y_2, ..., y_{N_s}]^T, y_i \in \mathcal{Y}_s$, the goal is to learn a classifier $f:\mathcal{X}\rightarrow \mathcal{Y}$. The classifier is usually used to recognize test data $\textbf{X}_u=[\textbf{x}_1^u, \textbf{x}_2^u, ..., \textbf{x}_{N_u}^u]$ from $\mathcal{Y}_u\subset \mathcal{Y}$ which is unseen during training, \textit{i.e.}, $\mathcal{Y}_s\cap\mathcal{Y}_u=\emptyset$.
%The true risk of $f$ w.r.t. unseen class data distribution $P_{\mathcal{X}_u,\mathcal{Y}_u}$ is $RI(f|P_{\mathcal{X}_u,\mathcal{Y}_u})=E_{P_{\mathcal{X}_u,\mathcal{Y}_u}}[\ell(f(\textbf{x}), y)]$, where $\ell:\mathcal{Y}\times\mathcal{Y}\rightarrow \mathbb{R}^+$ is a given loss function, for example the 0/1 error.

In the proposed Selective ZSC problem, the learner should output a selective classifier defined to be a pair $(f, g)$, where $f$ is a standard zero-shot classifier, and $g:\mathcal{X}\rightarrow\{0, 1\}$ is a selection function which is usually defined as $g(\textbf{x}) = \mathbbm{1}\{conf(\textbf{x})>\tau\}$. $conf$ is a confidence function, $\tau$ is a confidence threshold, and $\mathbbm{1}$ is an indicator function.
Given a test sample $\textbf{x}$,
\begin{equation}
(f, g)(\textbf{x}) \triangleq
\left\{
             \begin{array}{lr}
             f(\textbf{x}), &g(\textbf{x})=1\\
             reject, &g(\textbf{x})=0\\
             \end{array}
\right.
\end{equation}
The selective zero-shot classifier abstains from prediction when $g(\textbf{x})=0$. Its performance is usually evaluated by the risk-coverage curve~\cite{El-Yaniv:2010:FNS:1756006.1859904,NIPS2017_7073}. More details about the evaluation metric can be found in Section~\ref{sec:exp:data_and_settings}.

\section{The Proposed Selective Zero-Shot Classifier}

%We formulate the augmented attributes learning and the category prediction in a unified model where the classification task can be directly optimized.
In this section, we assume the model for augmented attributes has been learned and introduce our proposed selective classifier ($f, g$) based on the augmented attributes. Then in the next section, we introduce how the augmented attributes are learned.

Let $\mathcal{D}$ be the defined attribute space and $\mathcal{R}$ be the residual attribute space. For each $\textbf{x}\in\mathcal{X}$, we can obtain its augmented attribute prediction $[\textbf{d}; \textbf{r}]\in\mathcal{D}\mathcal{R}$ by the trained attribute model, where $\mathcal{D}\mathcal{R}=\mathcal{D}\times\mathcal{R}$. In zero-shot learning, for each seen category $y_s\in\mathcal{Y}_s$, an attribute annotation $\textbf{d}^{y_s}$ of the defined attributes is given. $\textbf{D}_o\in\mathbb{R}^{K_d\times|\mathcal{Y}_s|}$ is the class-level attribute annotation matrix, where the $i$-th column vector denotes the defined attribute annotation for the $i$-th seen category. Since no annotations of residual attributes are provided for the seen categories, we adopt the center of residual attribute predictions for each seen category as its residual attribute representation, denoted by $\textbf{r}^{y_s}$. Let $\textbf{R}_o\in\mathbb{R}^{K_r\times|\mathcal{Y}_s|}$ be the class-level residual attribute representation matrix. During the test phase, only the defined attributes are annotated for unseen categories ($\textbf{d}^{y_u}$ for $y_u \in \mathcal{Y}_u$).

\subsection{Zero-Shot Classifier $f$}
The zero-shot classifier $f$ is built on the defined attributes solely, as no annotations for residual attributes are provided. Given the defined attribute prediction $\hat{\textbf{d}}$ of a test image, the classifier $f$ is constructed by some form of nearest neighbor search
\begin{equation}
\hat{y} = \arg\max_{k\in{\mathcal{Y}_u}}{sim(\hat{\textbf{d}}, \textbf{d}^{k})},
\end{equation}
where $sim$ is the similarity function. In fact, many ZSC approaches follow the above general formulation, even though they may differ in the concrete form of $sim$. In this paper, it is simply defined as the cosine similarity.

\begin{figure}[t]
  \centering
  \includegraphics [scale=0.35]{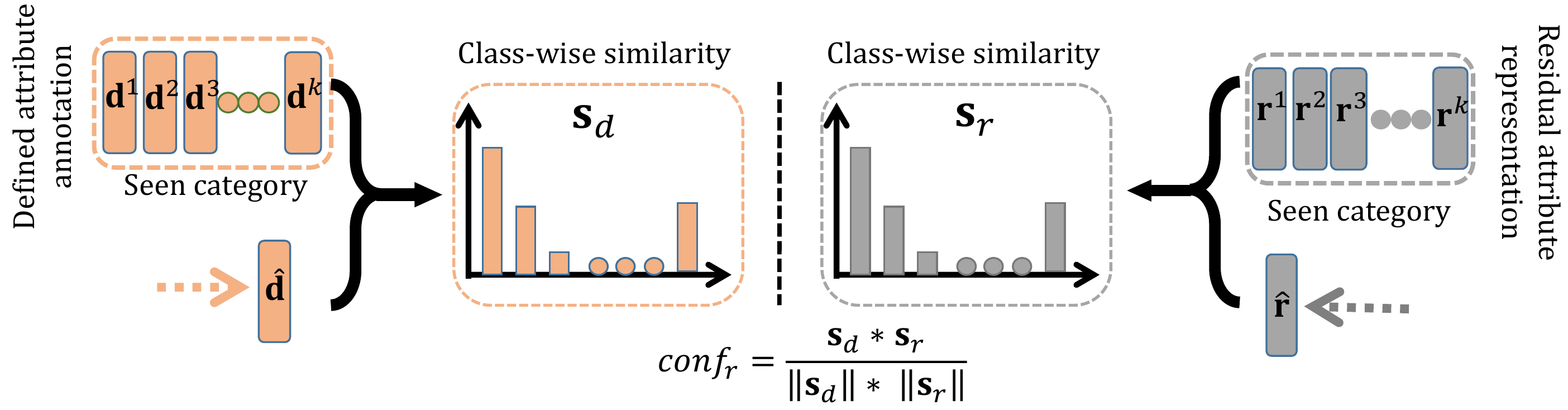}
  \caption{The confidence defined with the aid of the residual attributes.}
  \label{fig:consistency}
\end{figure}

\subsection{Confidence Function}

With $sim(\cdot)$ defined within the subspace of the manually defined attributes, the prediction confidence can be defined as the similarity score:
\begin{equation}
conf_d = sim(\hat{\textbf{d}}, \textbf{d}^{k}).
\end{equation}
However, as aforementioned, the defined attribute vocabulary alone is limited in its discriminative power. Thus the confidence score obtained within the defined attribute subspace is shortsighted. To tackle this issue, we propose to explore and exploit the residual attributes to overcome the shortcomings of the confidence produced by the defined attributes. Fig.~\ref{fig:consistency} illustrates the confidence score produced resorting to the residual attributes. Specifically, given a test image from an unseen class, we can obtain its augmented attribute presentation ($[\hat{\textbf{d}}; \hat{\textbf{r}}]$) by feeding the test image to the attribute prediction model. With this attribute presentation, two similarity vectors ($\textbf{s}_d$, $\textbf{s}_r$) can be computed: $\textbf{s}_d$ for the defined attributes and $\textbf{s}_r$ for the residual attributes. In these similarity vectors, the value of dimension $k$ measures the similarity between the predicted attributes and attribute presentation of class $k$. We formulate the similarity vector learning task as a sparse coding problem:
%, as similarly done in~\cite{zhang2015zero}:

\begin{equation}
\textbf{s}_d = \arg\min_{\textbf{s}}{\left\{\frac{\gamma}{2}\left\|\textbf{s}\right\|^2+\frac{1}{2}\left\|\hat{\textbf{d}}-\textbf{D}_o\textbf{s}\right\|_F^2\right\}},
\end{equation}

\begin{equation}
\textbf{s}_r = \arg\min_{\textbf{s}}{\left\{\frac{\gamma}{2}\left\|\textbf{s}\right\|^2+\frac{1}{2}\left\|\hat{\textbf{r}}-\textbf{R}_o\textbf{s}\right\|_F^2\right\}}.
\end{equation}
Then the confidence score can then be defined as the consistency of these two vectors:

\begin{equation}
conf_r = sim(\textbf{s}_d, \textbf{s}_r).
\end{equation}
The above confidence function is built on the intuition that the more consistent the defined and the residual attributes are, the less additional discriminative information the residual attributes provide for the current test image. Therefore, classification based on the defined attributes solely approximates classification based on the whole augmented attributes. Imagine that the residual attributes produce the same similarity vector as the defined attributes, then the residual attributes completely agree with the defined attribute on the prediction they made. However, if the residual attributes produce absolutely different similarity vector, then they do not reach a consensus. The defined attributes are shortsighted in this case and the produced prediction is more unreliable.

Combining the confidence function defined within the defined attribute subspace and that defined with the aid of residual attributes, the final confidence is
\begin{equation}
\label{eq:confidence}
conf = (1 - \lambda)conf_{d} + \lambda conf_{r},
\end{equation}
where $\lambda$ is a trade-off hyper-parameter which is set via cross-validation.

\section{Augmented Attribute Learning}

In this section, we introduce how the augmented attributes are learned. We formulate the augmented attribute learning task as a dictionary learning problem. The dictionary space is decomposed into two parts: (1) $\textbf{Q}_d$ corresponding to the defined-attribute-correlated dictionary subspace part which is correlated to the defined attribute annotations and the class annotations,  (2) $\textbf{Q}_r$ corresponding to the residual attribute dictionary subspace which is correlated to the class annotations and thus also useful for the classification task. To learn the whole dictionary space, three criteria are incorporated: (1) the defined attributes alone should be able to accomplish the classification task as better as possible; (2) the residual attributes should complement the discriminative power of defined attributes for classification; (3) the residual attributes should not rediscover the patterns that exist in the defined attributes. With all the three criteria, the objective function is formulated as:
\begin{equation}
\label{eq:objective_func}
\begin{split}
&\arg\min_{\{\textbf{Q}_d, \textbf{L}, \textbf{Q}_l, \textbf{U}, \textbf{Q}_r, \textbf{R}_s, \textbf{V}\}} \\
&\left\|\textbf{X}_s-\textbf{Q}_d\textbf{L}\right\|_F^2 + \alpha\left\|\textbf{L}-\textbf{Q}_l\textbf{D}_s\right\|_F^2 + \beta\left\|\textbf{H}-\textbf{U}\textbf{L}\right\|_F^2 \\
&+\left\|\textbf{X}_s-\textbf{Q}_d\textbf{L}-\textbf{Q}_r\textbf{R}_s\right\|_F^2 + \delta\left\|\textbf{H}-\textbf{U}\textbf{L}-\textbf{V}\textbf{R}_s\right\|_F^2\\
&-\eta\left\|\textbf{R}_s-\textbf{W}\textbf{L}\right\|_F^2,\\
&s.t. \ \textbf{W}=\arg\min_\textbf{W}{\left\|\textbf{R}_s-\textbf{W}\textbf{L}\right\|_2^2}, \ \left\|\textbf{w}_i\right\|_2^2\leq 1, \ \left\|\textbf{q}_{di}\right\|_2^2\leq 1, \ \\ &\left\|\textbf{q}_{ri}\right\|_2^2\leq 1, \left\|\textbf{q}_{li}\right\|_2^2\leq 1, \ \left\|\textbf{u}_i\right\|_2^2\leq 1, \ \left\|\textbf{v}_i\right\|_2^2\leq 1,\ \forall i.\\
\end{split}
\end{equation}
In the above formulation, the second, the third, and the fourth lines are corresponding to the first, the second, and the third criteria, respectively. As the proposed classifier $f$ makes predictions based on only the defined attributes, the first criterion protects $f$ from being distracted from its classification task. However, defined attributes are usually not equally valuable for classification and some of them are highly correlated. Instead of adopting the defined attributes directly, we employ discriminative latent attributes proposed in~\cite{Jiang_2017_ICCV} for zero-shot classification. $\textbf{L}$ is latent attributes which are derived from the defined attributes and $\textbf{H} = [\textbf{h}_1, \textbf{h}_2, ...]$ where $\textbf{h}_i = [0,...,0, 1, 0,...,0]^T$ is a one hot vector which gives the label of sample $i$. Thus $\textbf{U}$ can be regarded as the seen-class classifier in the latent attribute space. For the second criterion, we assume the learned residual attributes suffer little of the above problem and adopt them and the discriminative latent attributes jointly for the classification task. For the third criterion, as we expect the residual attributes discover non-redundant properties, the defined attributes should not be predictive for the residual attributes. $\textbf{w}_i$ is the $i$-th column of $\textbf{W}$.

Optimizing the three criteria simultaneously  is challenging as there are several hyper-parameters which are set via cross-validation. Furthermore, it may degrade the performance of $f$, as $f$ makes predictions based on the defined attributes solely. We divide the optimization problem in Eqn.~\ref{eq:objective_func} into two subproblems which are optimized separately. In the first subproblem, only the first criterion is considered and we optimize $\textbf{Q}_d, \textbf{L}, \textbf{Q}_l$ and $\textbf{U}$ to strive for $f$ with higher performance. In the second subproblem, $\textbf{Q}_d, \textbf{L}, \textbf{Q}_l$ and $\textbf{U}$ are fixed and we optimize $\textbf{Q}_r, \textbf{R}_s$ and $\textbf{V}$ with taking the second and the third criteria into consideration. With our proposed optimization procedure, the cross validation work for hyper-parameters $\{\alpha, \beta, \delta, \eta\}$ is significantly reduced as $\{\alpha, \beta\}$ and $\{\delta, \eta\}$ are cross validated separately.
\subsubsection{The First Subproblem.}
Taking only the first criterion into consideration, Eqn.~\ref{eq:objective_func} is simplified to be
\begin{equation}
\label{eq:subproblem_1}
\begin{split}
&\arg\min_{\{\textbf{Q}_d, \textbf{L}, \textbf{Q}_l, \textbf{U}\}}
\left\|\textbf{X}_s-\textbf{Q}_d\textbf{L}\right\|_F^2 + \alpha\left\|\textbf{L}-\textbf{Q}_l\textbf{D}_s\right\|_F^2 + \beta\left\|\textbf{H}-\textbf{U}\textbf{L}\right\|_F^2, \\
&s.t. \ \left\|\textbf{q}_{di}\right\|_2^2\leq 1, \left\|\textbf{q}_{li}\right\|_2^2\leq 1, \ \left\|\textbf{u}_i\right\|_2^2\leq 1, \ \forall i.\\
\end{split}
\end{equation}
This is the problem proposed in~\cite{Jiang_2017_ICCV}. Eqn.~\ref{eq:subproblem_1} is not convex for $\textbf{Q}_d, \textbf{L}, \textbf{Q}_l$ and $\textbf{U}$ simultaneously, but it is convex for each of them separately. An alternating optimization method is adopted to solve it. Detailed optimization process can be found in~\cite{Jiang_2017_ICCV}.

\subsubsection{The Second Subproblem.}
After solving the first subproblem, $\textbf{Q}_d, \textbf{L}, \textbf{Q}_l$ and $\textbf{U}$ are fixed and Eqn.~\ref{eq:objective_func} is simplified to be
\begin{equation}
\label{eq:subproblem_2}
\begin{split}
&\arg\min_{\{\textbf{Q}_r, \textbf{R}_s, \textbf{V}\}}
\left\|\textbf{X}_s-\textbf{Q}_d\textbf{L}-\textbf{Q}_r\textbf{R}_s\right\|_F^2 + \delta\left\|\textbf{H}-\textbf{UL}-\textbf{V}\textbf{R}_s\right\|_F^2-\eta\left\|\textbf{R}_s-\textbf{WL}\right\|_F^2,\\
&s.t. \ \textbf{W}=\arg\min_\textbf{W}{\left\|\textbf{R}_s-\textbf{WL}\right\|_2^2}, \ \left\|\textbf{w}_i\right\|_2^2\leq 1, \left\|\textbf{q}_{ri}\right\|_2^2\leq 1, \ \left\|\textbf{v}_i\right\|_2^2\leq 1, \ \forall i.\\
\end{split}
\end{equation}
Similarly, $\textbf{Q}_r, \textbf{R}_s$ and $\textbf{V}$ are optimized by the alternate optimization method. The optimization process is briefly described as follows.

(1) Fix $\textbf{Q}_r, \textbf{V}$ and update $\textbf{R}_s$:
\begin{equation}
\arg\min_{\textbf{R}_s}{\left\|\tilde{\textbf{X}}-\tilde{\textbf{Q}}\textbf{R}_s\right\|_F^2},
\end{equation}
where
\begin{equation*}
\tilde{\textbf{X}} =
\left [
             \begin{array}{lr}
             \textbf{X}_s-\textbf{Q}_d\textbf{L}\\
             \delta(\textbf{H}-\textbf{UL})\\
             -\eta(\textbf{WL})\\
             \end{array}
\right], \ \tilde{\textbf{Q}}=
\left [
             \begin{array}{lr}
             \textbf{Q}_r\\
             \delta\textbf{V}\\
             -\eta \textbf{I}\\
             \end{array}
\right],
\end{equation*}
and $\textbf{I}$ is the identity matrix. $\textbf{R}_s$ has the closed-form solution as
\begin{equation}
\textbf{R}_s= (\tilde{\textbf{Q}}^T\tilde{\textbf{Q}})^{-1}\tilde{\textbf{Q}}^T\tilde{\textbf{X}}.
\label{eq:Rs}
\end{equation}

(2) Fix $\textbf{R}_s, \textbf{V}$ and update $\textbf{Q}_r$:
\begin{equation}
\arg\min_{\textbf{Q}_r}\left\|\textbf{X}_s-\textbf{Q}_d\textbf{L}-\textbf{Q}_r\textbf{R}_s\right\|_F^2, \ s.t. \ \left\|\textbf{q}_{ri}\right\|_2^2\leq 1, \ \forall i.
\label{eq:sec}
\end{equation}
The above problem can be solved by the Lagrange dual and the analytical solution is
\begin{equation}
\textbf{Q}_r= (\textbf{X}_s-\textbf{Q}_d\textbf{L}){\textbf{R}_s}^T({\textbf{R}_s}{\textbf{R}_s}^T+\boldsymbol\Lambda)^{-1},
\label{eq:Qr}
\end{equation}
where $\boldsymbol\Lambda$ is a diagonal matrix constructed by all the Lagrange dual variables.

(3) Fix $\textbf{R}_s, \textbf{Q}_r$ and update $\textbf{V}$:
\begin{equation}
\arg\min_{\textbf{V}}\left\|\textbf{H}-\textbf{UL}-\textbf{V}\textbf{R}_s\right\|_F^2, \ s.t. \ \left\|\textbf{v}_i\right\|_2^2\leq 1, \ \forall i.
\end{equation}
The above problem can be solved in the same way as Eqn.~\ref{eq:sec} and the solution is
\begin{equation}
\textbf{V}= (\textbf{H}-\textbf{UL}){\textbf{R}_s}^T({\textbf{R}_s}{\textbf{R}_s}^T+\boldsymbol\Lambda)^{-1}.
\label{eq:V}
\end{equation}

(4) Computing $\textbf{W}$:
\begin{equation}
\label{eq:W}
\arg\min_\textbf{W}{\left\|\textbf{R}_s-\textbf{WL}\right\|_F^2}, \ s.t. \ \left\|\textbf{w}_i\right\|_2^2\leq 1, \ \forall i.
\end{equation}
Similar to Eqn.~\ref{eq:Qr} and Eqn.~\ref{eq:V}, we can get the solution
\begin{equation}
\textbf{W}= \textbf{R}_s{\textbf{L}}^T({\textbf{L}}{\textbf{L}}^T+\boldsymbol\Lambda)^{-1}.
\label{eq:W}
\end{equation}

The complete algorithm is summarized in Algorithm~\ref{ag:attr_learning}. The optimization process usually converges quickly, after tens of iterations in our experiments.
\begin{algorithm}[h]
\caption{Augmented Attribute Learning for Selective ZSC}
{\bf Input}:
$\textbf{X}_s, \textbf{D}_s, \textbf{H}, \alpha, \beta, \delta, \eta, K_r$ \\
{\bf Output}:
$\textbf{Q}_d, \textbf{L}, \textbf{Q}_l, \textbf{U}, \textbf{Q}_r, \textbf{R}_s, \textbf{V}$
\begin{algorithmic}[1]
\State Optimizing the first subproblem according to~\cite{Jiang_2017_ICCV}, obtaining $\textbf{Q}_d, \textbf{L}, \textbf{Q}_l, \textbf{U}$.
\State Fixing $\textbf{Q}_d, \textbf{L}, \textbf{Q}_l, \textbf{U}$ obtained from 1, and initializing $\textbf{Q}_r, \textbf{V}, \textbf{W}$ randomly according to $K_r$.
\While{not converge}
\State Optimizing $\textbf{R}_s$ according to Eqn.~\ref{eq:Rs}.
\State Optimizing $\textbf{Q}_r$ according to Eqn.~\ref{eq:Qr}.
\State Optimizing $\textbf{V}$ according to Eqn.~\ref{eq:V}.
\State Optimizing $\textbf{W}$ according to Eqn.~\ref{eq:W}.
\EndWhile
\State \Return $\textbf{Q}_d, \textbf{L}, \textbf{Q}_l, \textbf{U}, \textbf{Q}_r, \textbf{R}_s, \textbf{V}$
\end{algorithmic}
\label{ag:attr_learning}
\end{algorithm}
\section{Experiments}
\subsection{Datasets and Settings}
\label{sec:exp:data_and_settings}
\subsubsection{Datasets.}
We conduct experiments on three benchmark image datasets for ZSC, including aPascal$\&$aYahoo (\textbf{aP$\&$Y})~\cite{farhadi2009describing}, Animals with Attributes (\textbf{AwA})~\cite{lampert2009learning} and Caltech-UCSD Birds-200-2011 (\textbf{CUB-200})~\cite{WelinderEtal2010}. For all the datasets, we split the categories into seen and unseen sets in the same way as~\cite{Jiang_2017_ICCV}: (1) There are two attribute datasets in \textbf{aP$\&$Y}: aPascal and aYahoo. These two datasets contains images from disjoint object classes. The categories in aPascal dataset are used as seen classes and those in aYahoo as the unseen ones. (2)~\textbf{AwA} contains 50 categories, 40 of which are used as seen categories, and the rest 10 are used as the unseen ones. (3) \textbf{CUB-200} is a bird dataset for fine-grained recognition. It contains 200 categories, of which 150 are used as seen categories and the rest 50 as the unseen ones. For all the datasets, we adopt the pre-trained VGG19~\cite{simonyan2014very} to extract features.

\subsubsection{Cross validation.} There are several hyper-parameters (including $\gamma, K_r, \alpha, \beta, \delta, \eta$) which are set via cross-validation. As aforementioned, our proposed optimization procedure relaxes the laborious cross-validation work by decomposing the original problem into two subproblems. $\alpha, \beta$ are firstly optimized on the validation data independent of the others. After that, to further relax the cross-validation work, we optimize $\delta, \eta, K_r$ independent of $\gamma$. Finally, $\gamma$ is optimized. In this paper, we adopt five-fold cross-validation~\cite{zhang2015zero} for all these parameters.

\subsubsection{Evaluation Metrics.}
The performance of the classifier is quantified using \textit{coverage} and \textit{risk}. The coverage is defined to be the probability mass of the non-rejected region in $\mathcal{X}_u$ (the feature space of unseen classes)

\begin{equation}
coverage(f, g) \triangleq E_p[g(\textbf{x})],
\end{equation}
and the selective risk of $(f, g)$ is

\begin{equation}
risk(f, g) \triangleq \frac{E_p[\ell(f(\textbf{x}), y)g(\textbf{x})]}{coverage(f, g)},
\end{equation}
where $\ell$ is defined to be $0/1$ loss. The risk can be traded off for coverage. Thus the overall performance of a selective classifier can be measured by its Risk-Coverage Curve (RCC),  where risk is defined to be a function of the coverage~\cite{El-Yaniv:2010:FNS:1756006.1859904,NIPS2017_7073}. The Area Under Risk-Coverage Curve (AURCC) is usually adopted to quantify the performance.

\subsection{Ablation Study}

\subsubsection{The Effectiveness of Three Criteria.}
\begin{figure}[t]
  \centering
  \includegraphics [scale=0.20]{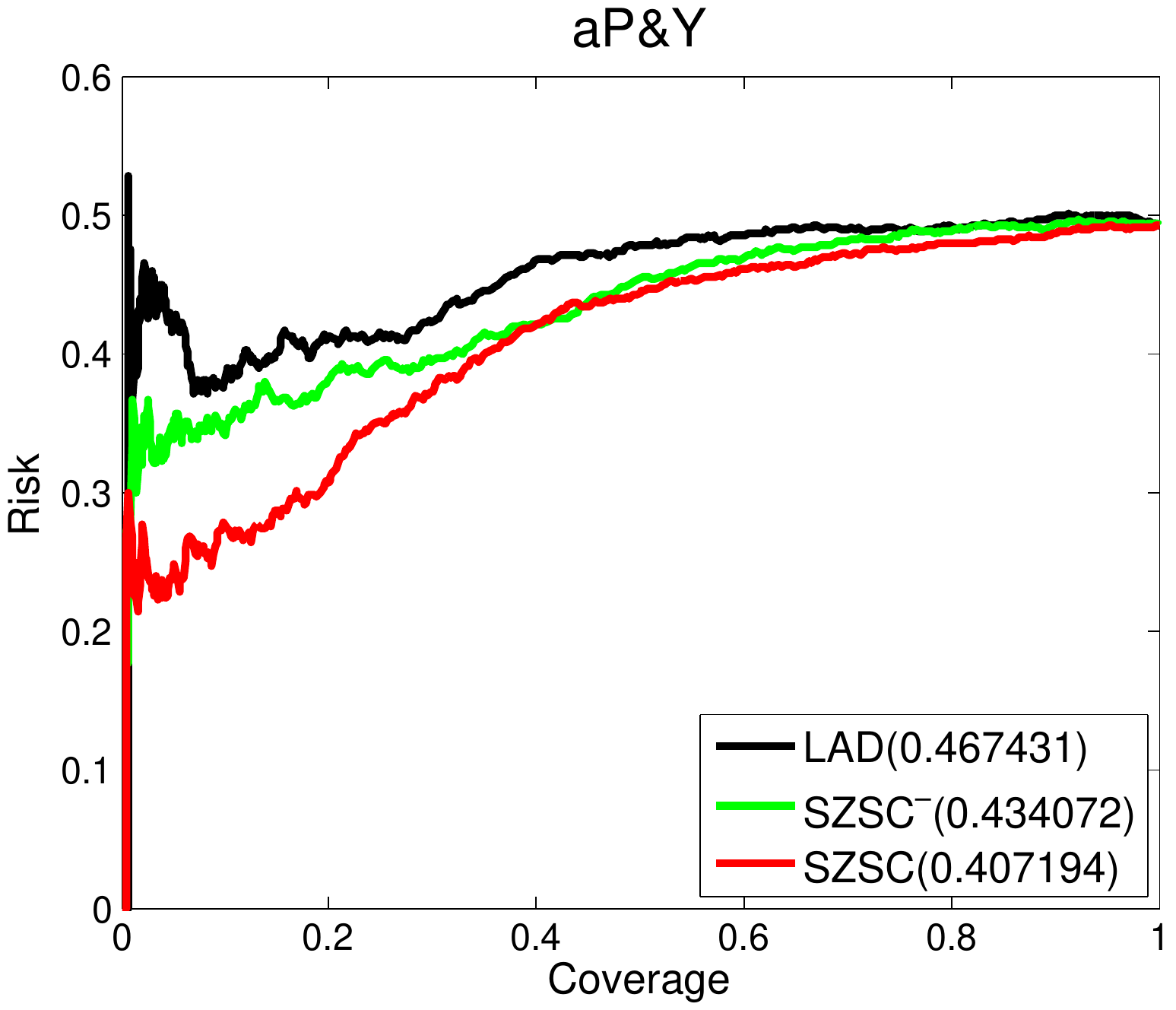}
  \includegraphics [scale=0.20]{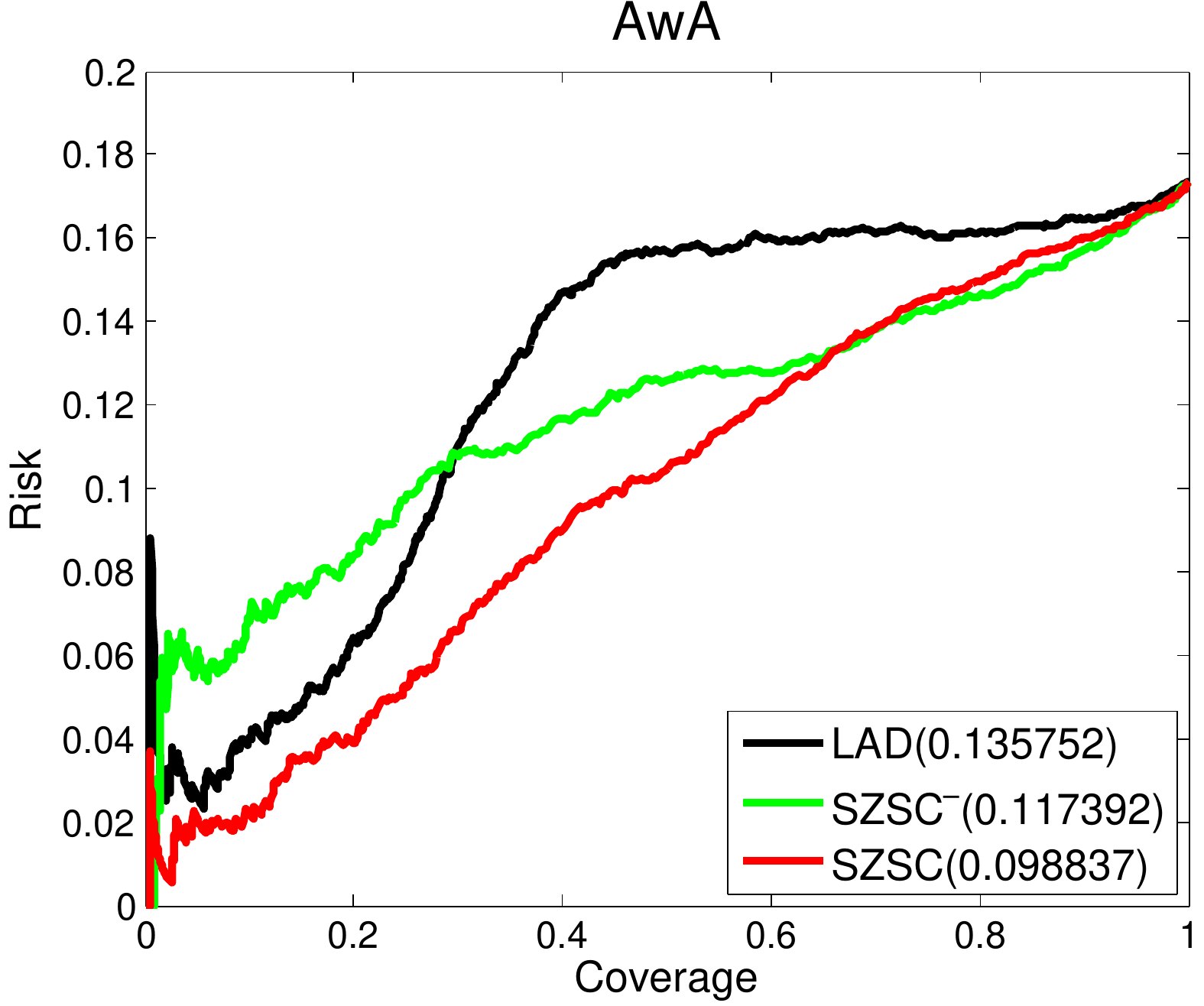}
  \includegraphics [scale=0.20]{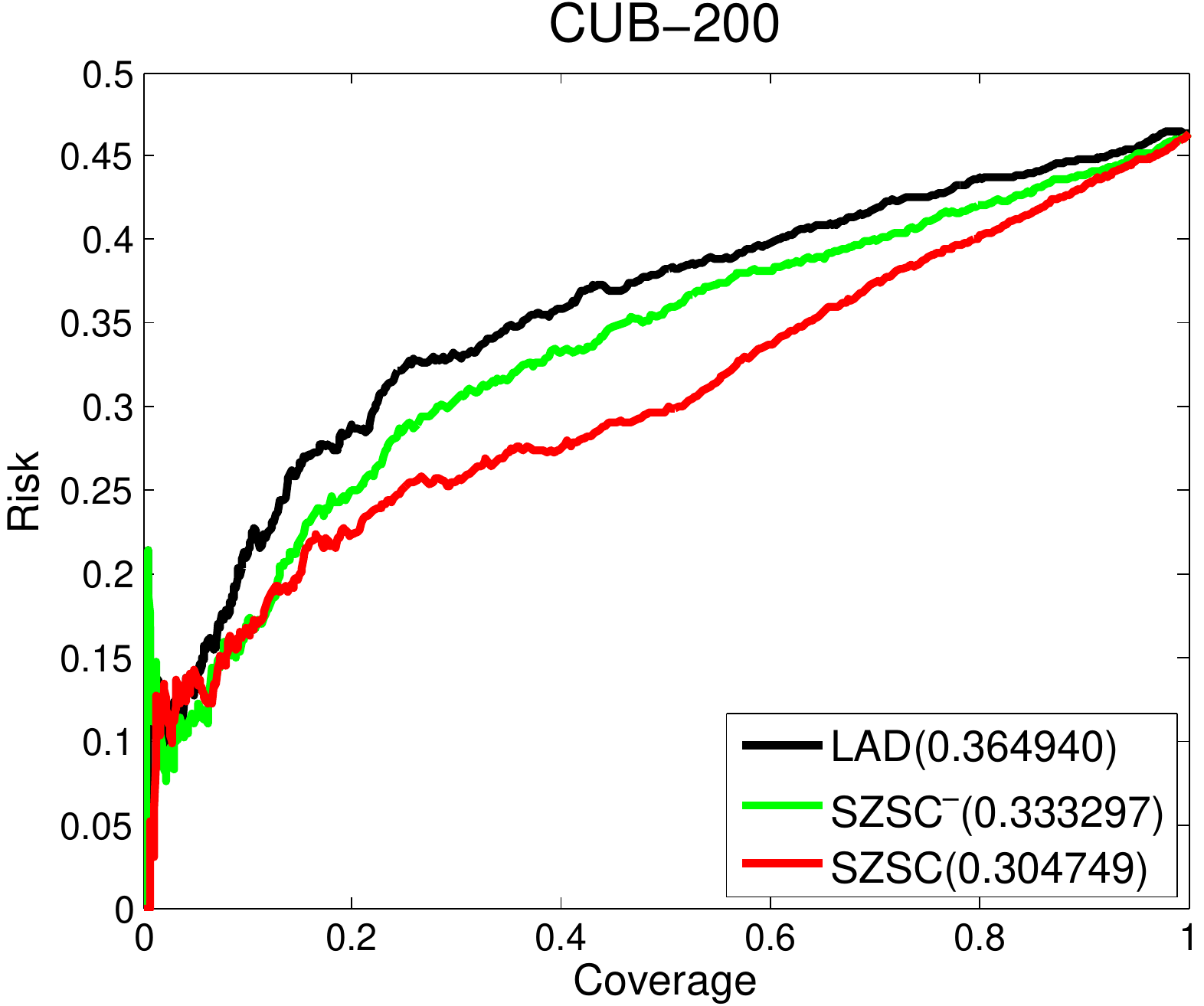}
  \caption{Comparisons among different variants of the proposed method (best viewed in color). AURCC is given in brackets.}
  \label{fig:ablation_criteria}
\end{figure}

We have incorporated three criteria into the learning of augmented attributes. In this section, we validate the effectiveness of them. We make comparisons among three variants of the proposed method. For the first one, only the first criterion is considered. In other words, no residual attributes are learned, and the classification model degrades to LAD~\cite{Jiang_2017_ICCV} ($conf = conf_d$). For the second one (dubbed SZSC$^-$), the first and the second criteria are considered. For the third one (dubbed SZSC), all the three criteria are incorporated. For all the three variants, the dimensions of the residual attributes are kept the same as that of the defined attributes ($K_r = K_d$). Other hyper-parameters are set via cross-validation. The risk-coverage curves on all the three benchmark datasets are depicted in Fig.~\ref{fig:ablation_criteria}. It can be seen that on all the three datasets, the proposed method achieves the best performance when all the three criteria are involved.

\subsubsection{Trade-off between Two Confidence Scores.}
\begin{figure}[t]
  \centering
  \includegraphics [scale=0.20]{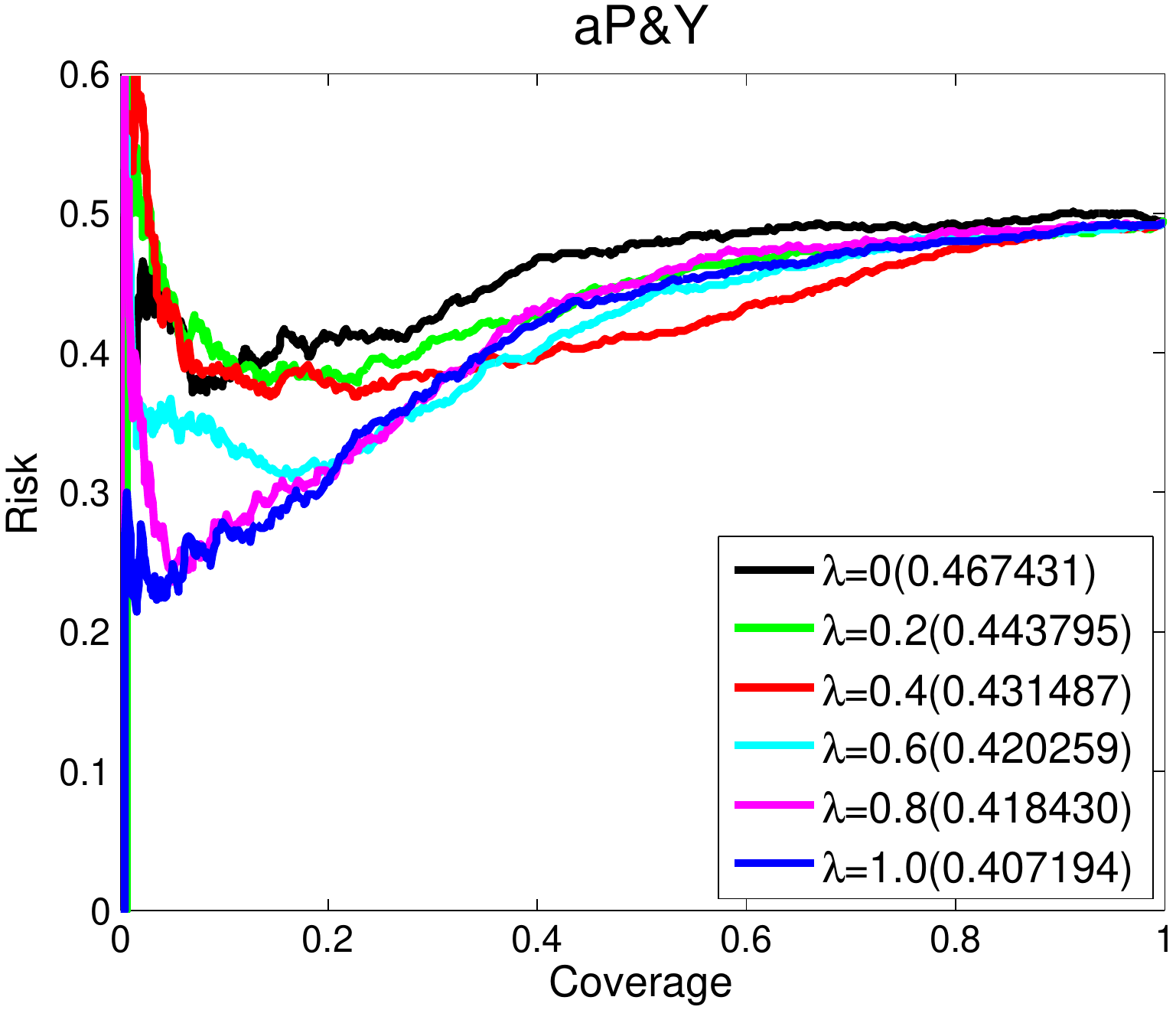}
  \includegraphics [scale=0.20]{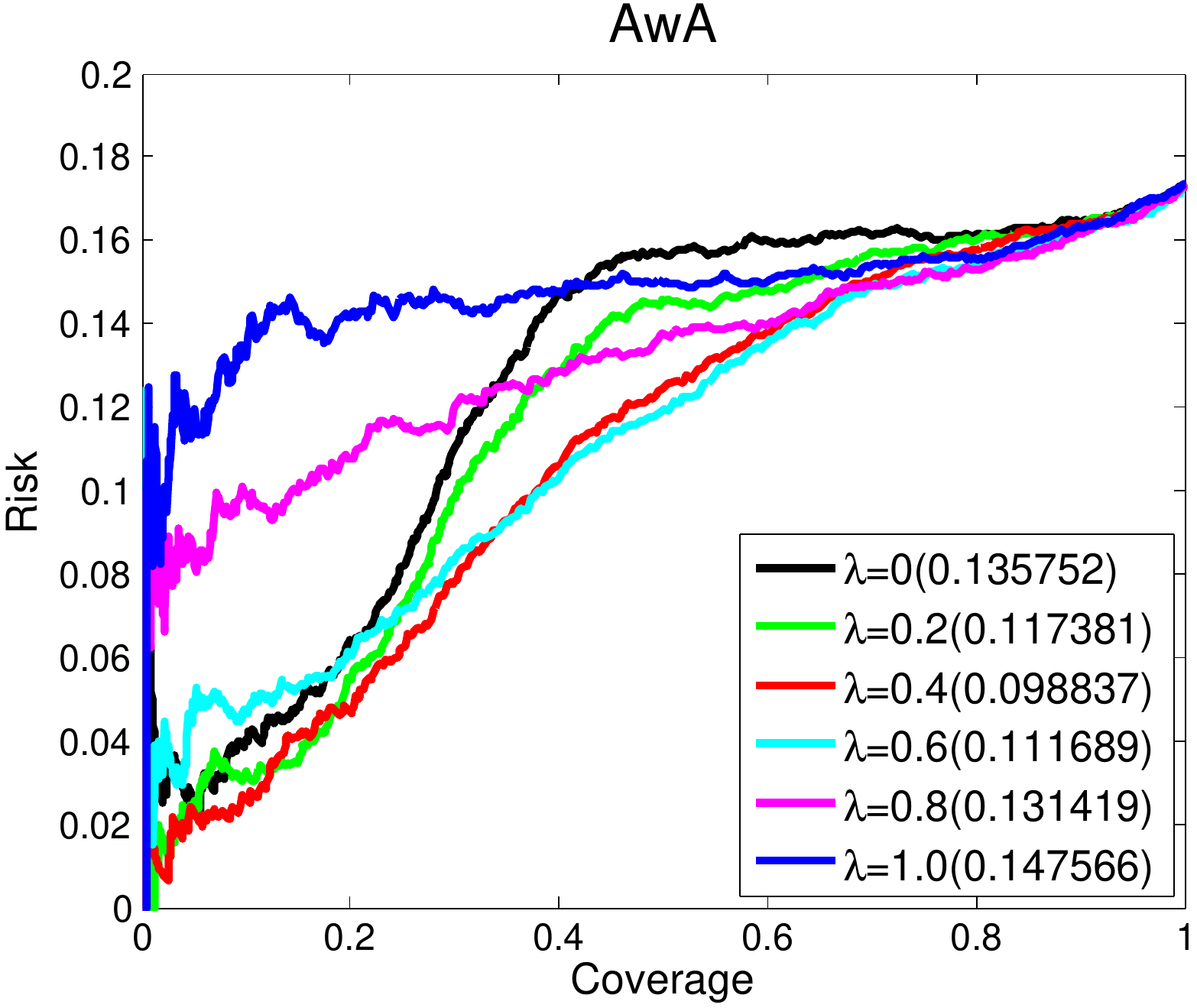}
  \includegraphics [scale=0.20]{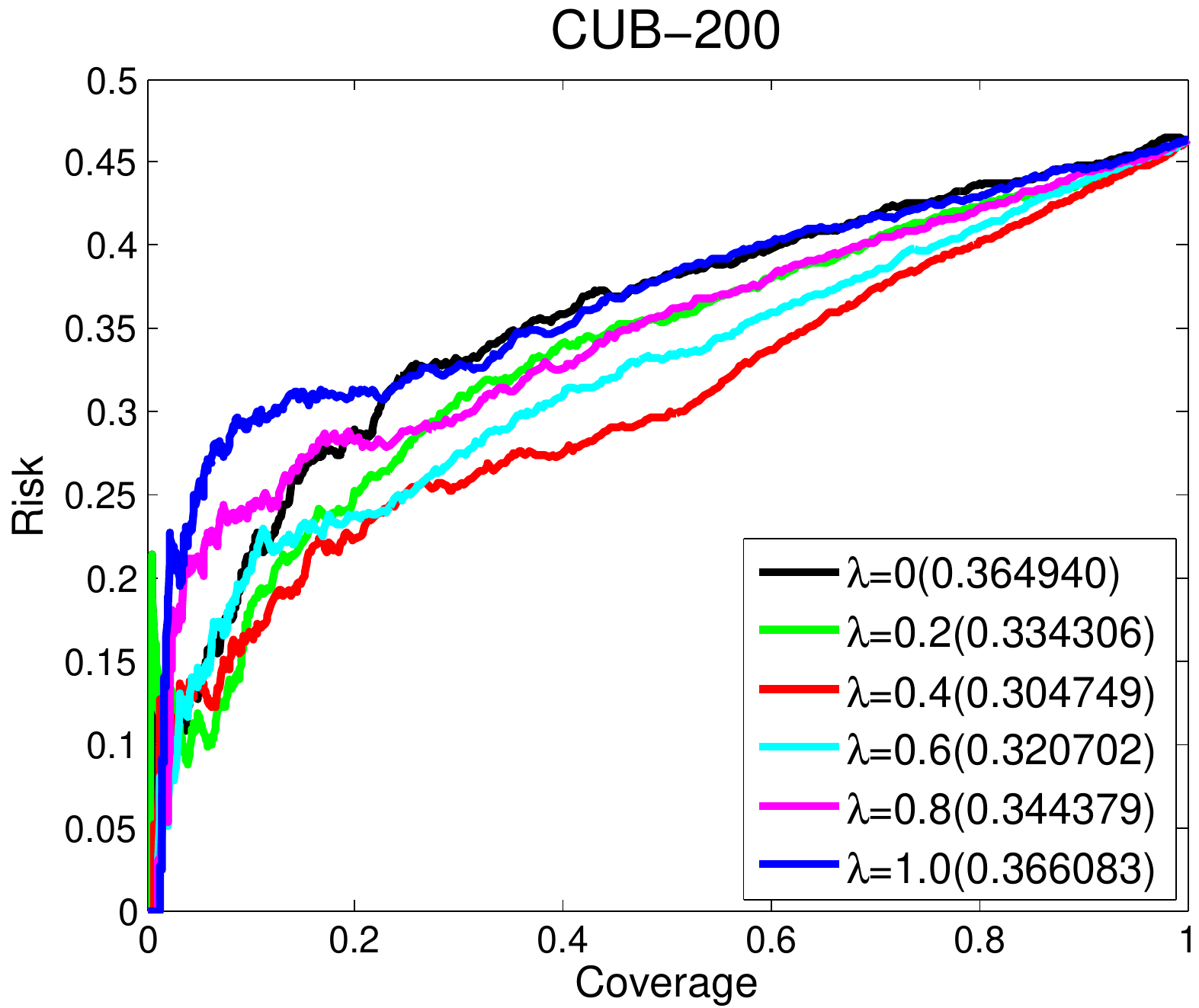}
  \caption{Risk-coverage curves of the proposed method with varying $\lambda$.}
  \label{fig:ablation_trade_off}
\end{figure}

The proposed confidence function is composed of two parts: the confidence defined within the defined attributes ($conf_d$) and the confidence defined with the aid of the residual attributes ($conf_r$). In this section, we test how the trade-off parameter $\lambda $ affects the performance of SZSC. If $\lambda=0$, the confidence depends entirely on the defined attributes. On the contrary, if $\lambda=1$, the confidence is composed of $conf_r$ only. All other hyper-parameters are kept the same for fair comparisons. Experimental results on all the three benchmark datasets are shown in Fig.~\ref{fig:ablation_trade_off}. It reveals that the appropriately combined confidence significantly improve the classifier's performance on all the three datasets. More surprisingly, on aP\&Y the optimally combined confidence relies heavily on $conf_r$ ($\lambda=1.0$). The under-complete defined attribute vocabulary and the large difference between the seen and the unseen categories may account for that.

\subsubsection{Dimensions of the Residual Attribute Space.}

\begin{figure}[t]
  \centering
  \includegraphics [scale=0.20]{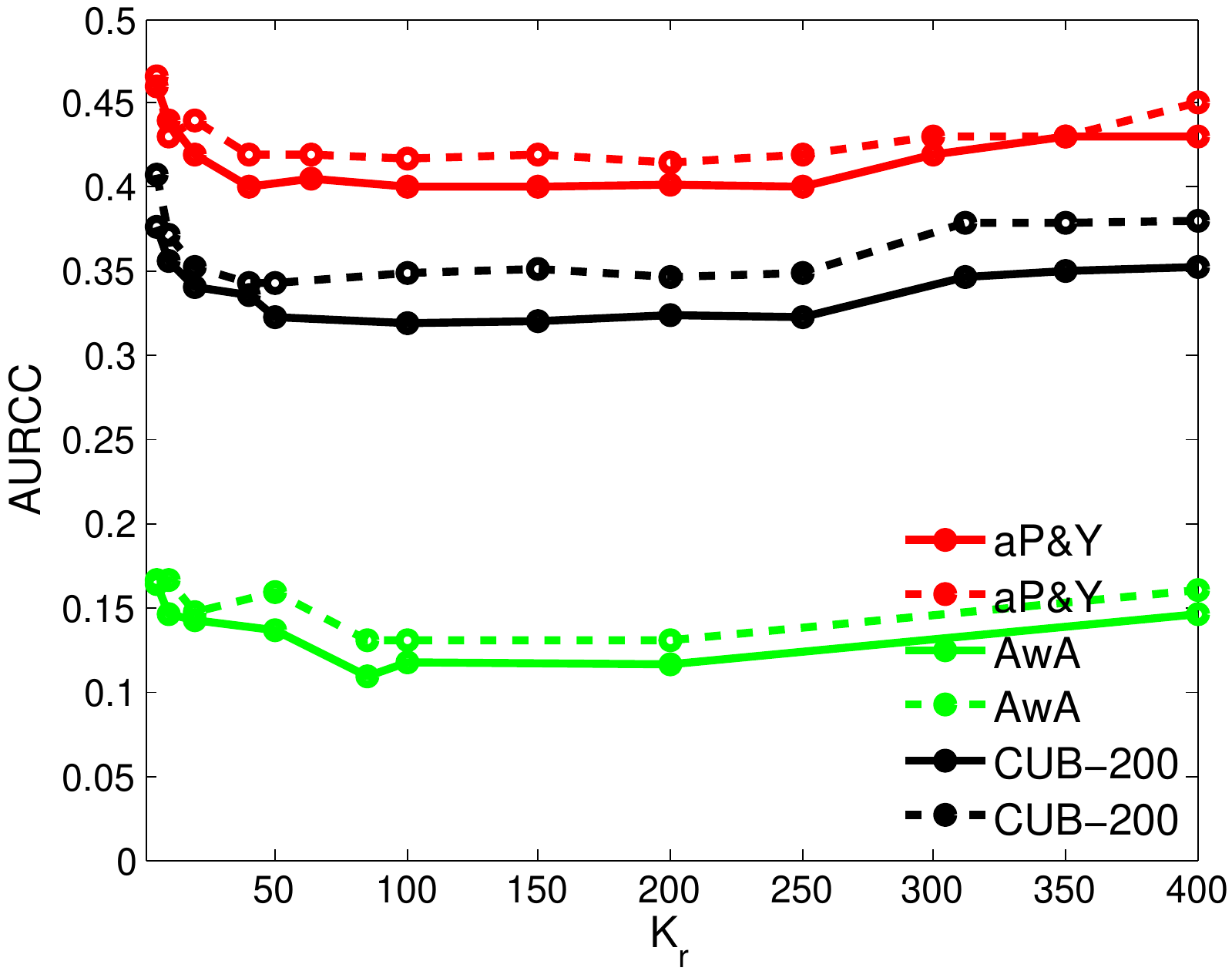}
  \includegraphics [scale=0.20]{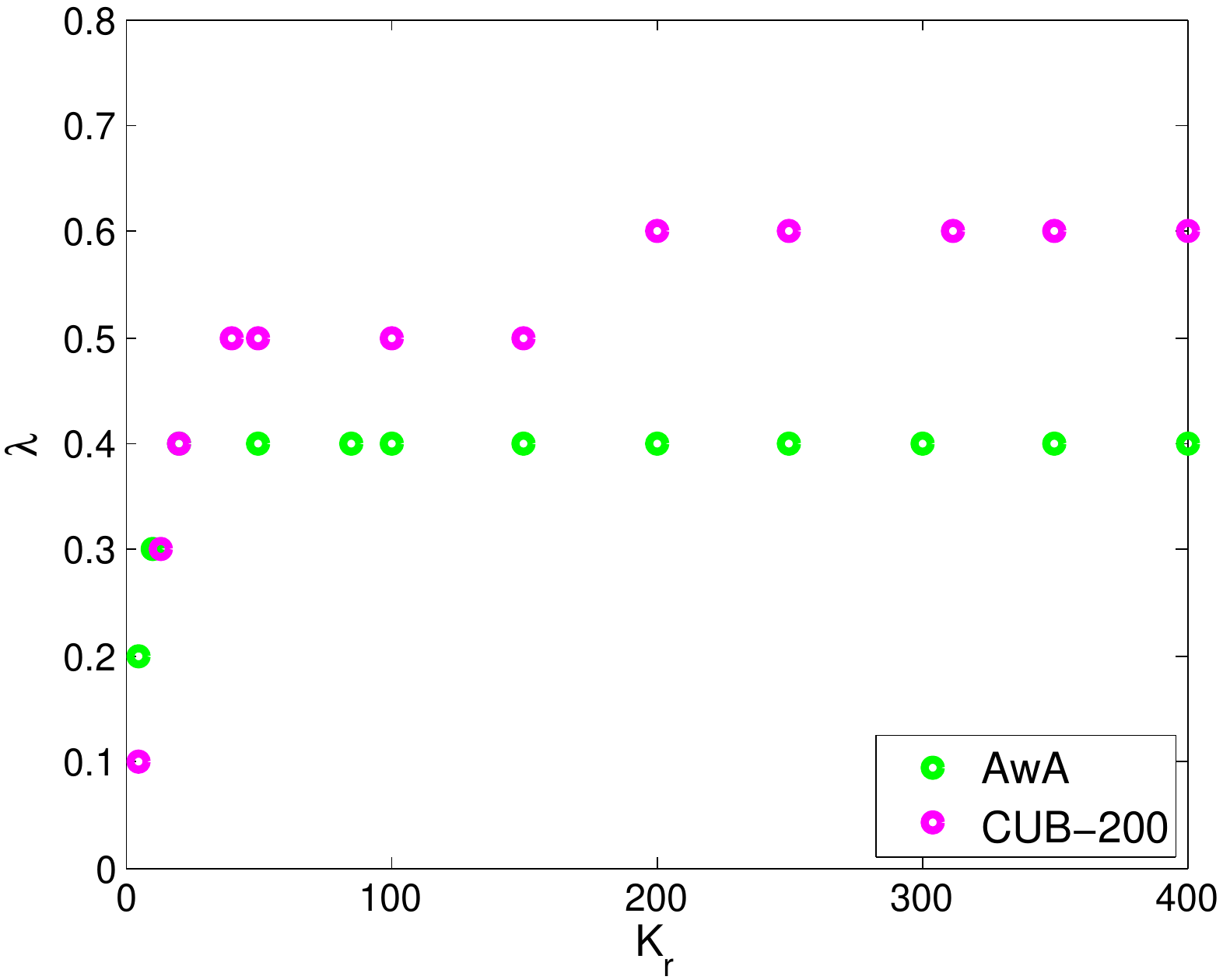}
  \caption{How AURCC (left) and the optimal $\lambda$ (right) change with varying $K_r$.}
  \label{fig:ablation_dimension}
\end{figure}

In this section, we investigate how the performance changes with the varying $K_r$. Similarly, all other hyper-parameters are kept the same. For a more comprehensive view of the proposed method, both SZSC$^-$ and SZSC are evaluated. Experimental results are shown in Fig.~\ref{fig:ablation_dimension} (left). It can be observed that the number of dimensions of the residual attribute space also makes unneglected impacts on the final performance. Too small $K_r\ (< 50)$ will leave the residual discriminative properties not fully explored. Conversely, too large $K_r\ (>300)$ renders the optimization more challenging and time-consuming. Both these two cases degrade the performance. Furthermore, we test that how the cross-validated $\lambda$ changes with $K_r$. Results are depicted in Fig.~\ref{fig:ablation_dimension} (right). It can be seen that with small $K_r$, the confidence obtained via the residual attributes is unreliable and the optimally combined confidence relies heavily (small $\lambda$) on the $conf_d$. However, as $K_r$ becomes larger, $\lambda$ also becomes larger which indicates that $conf_r$ plays a more important role.

\subsection{Benchmark Comparison}
\label{ex:benchmark}
\subsubsection{Competitors.}
Several existing ZSC models are selected for benchmark comparison, including SSE~\cite{zhang2015zero}, SYNC~\cite{changpinyo2016synthesized}, SCoRe~\cite{Morgado_2017_CVPR}, SAE~\cite{Kodirov_2017_CVPR} and LAD~\cite{Jiang_2017_ICCV}. The selection criteria are (1) representativeness: they cover a wide range of models; (2) competitiveness: they clearly represent the state-of-the-art; (3) recent work: all of them are published in the past three years; (4) reproducibility: all of them are code available, so the provided results in this paper are reproducible. We briefly review them and introduce their typical confidence functions as follows. SSE adopts SVM as the classification model. The margin in SVM classifier is employed as the confidence. SCoRe utilizes deep neural networks integrated with a softmax classifier for ZSC. The softmax output is usually employed for misclassification or out-of-distribution example dection~\cite{hendrycks2016baseline}. We also use it as the proxy of the confidence for SCoRe. For the other competitors, the classification task is usually accomplished via nearest neighbor searches in the shared embedding space. We take the cosine similarity as the confidence.

\begin{figure}[t]
  \centering
  \includegraphics [scale=0.20]{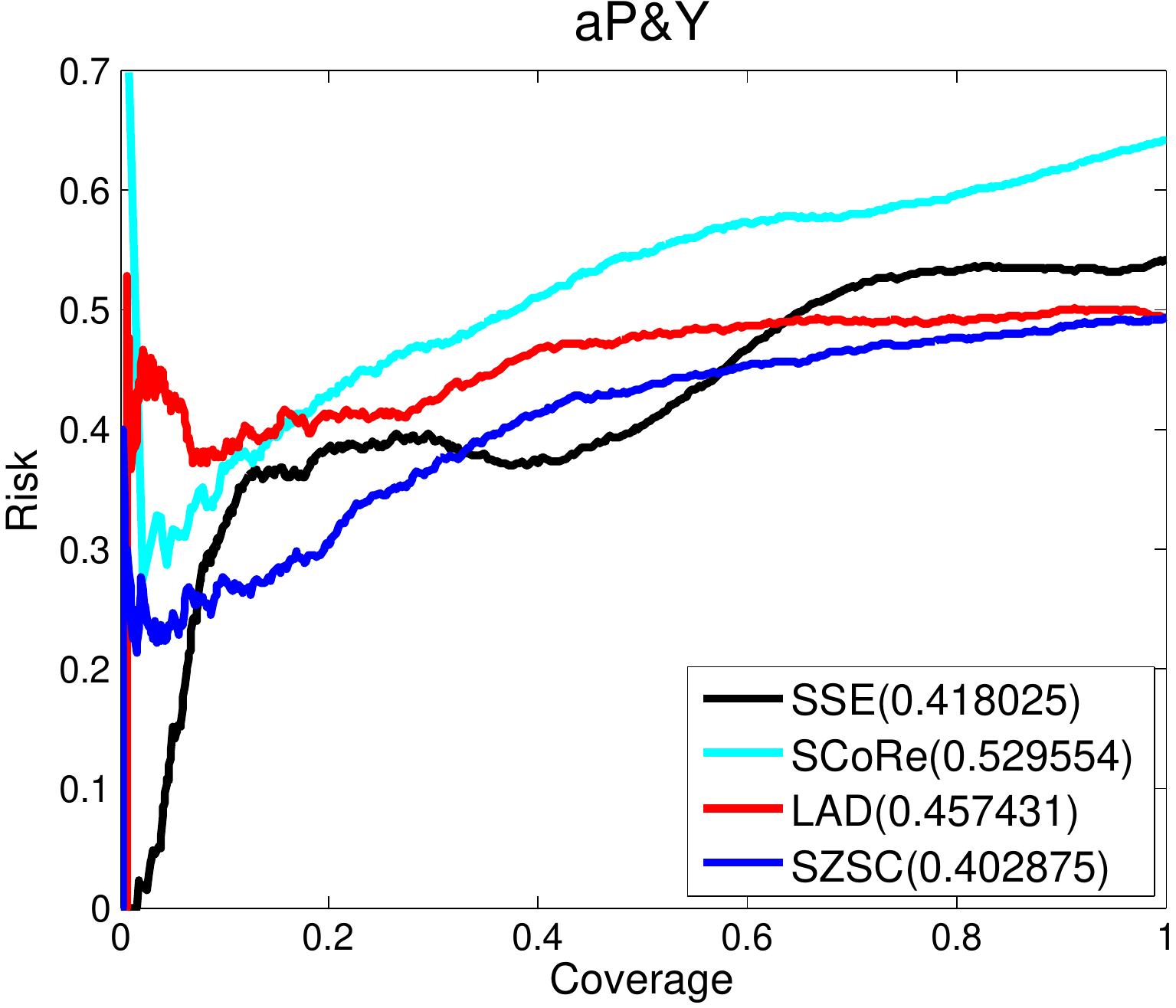}
  \includegraphics [scale=0.20]{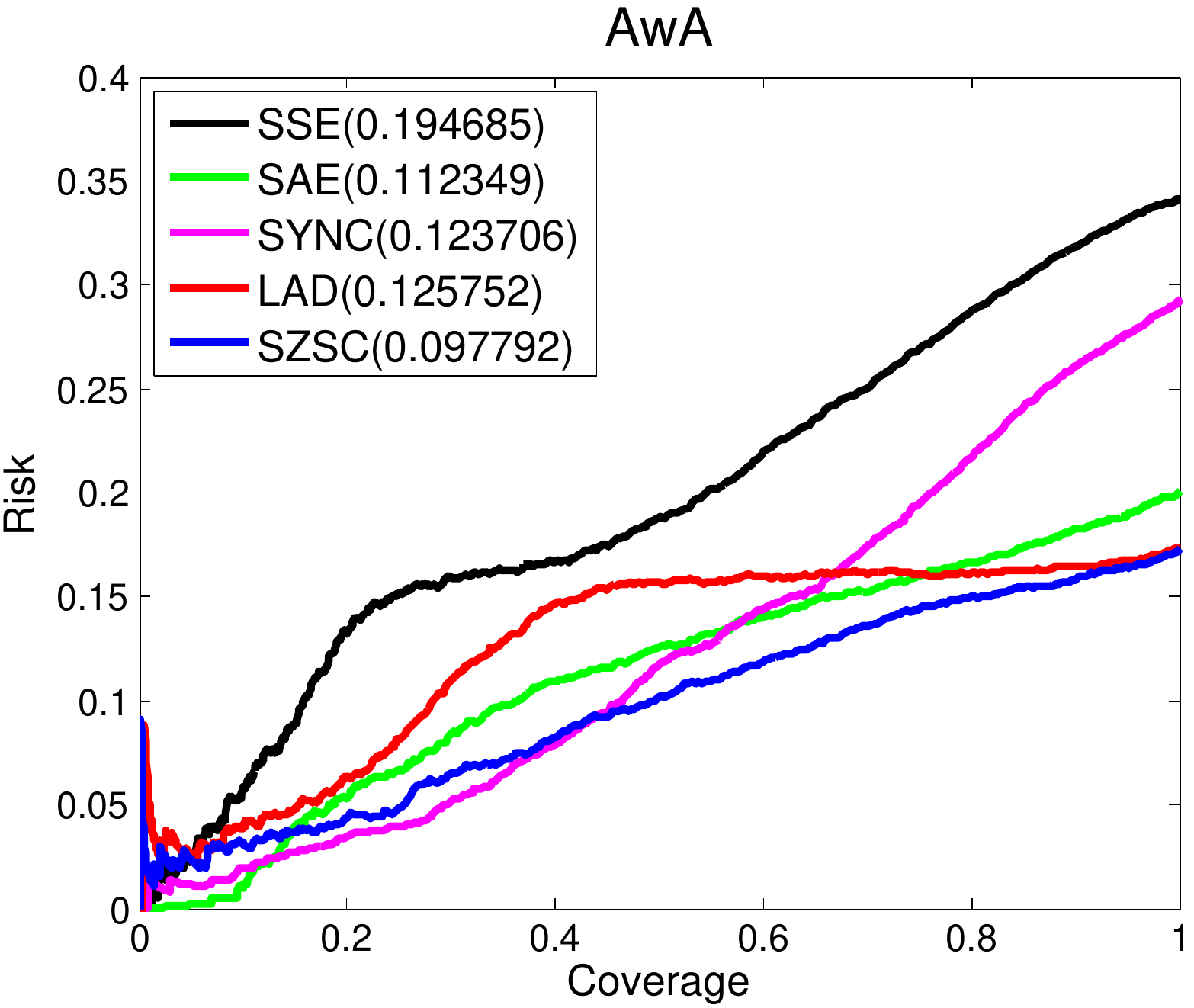}
  \includegraphics [scale=0.20]{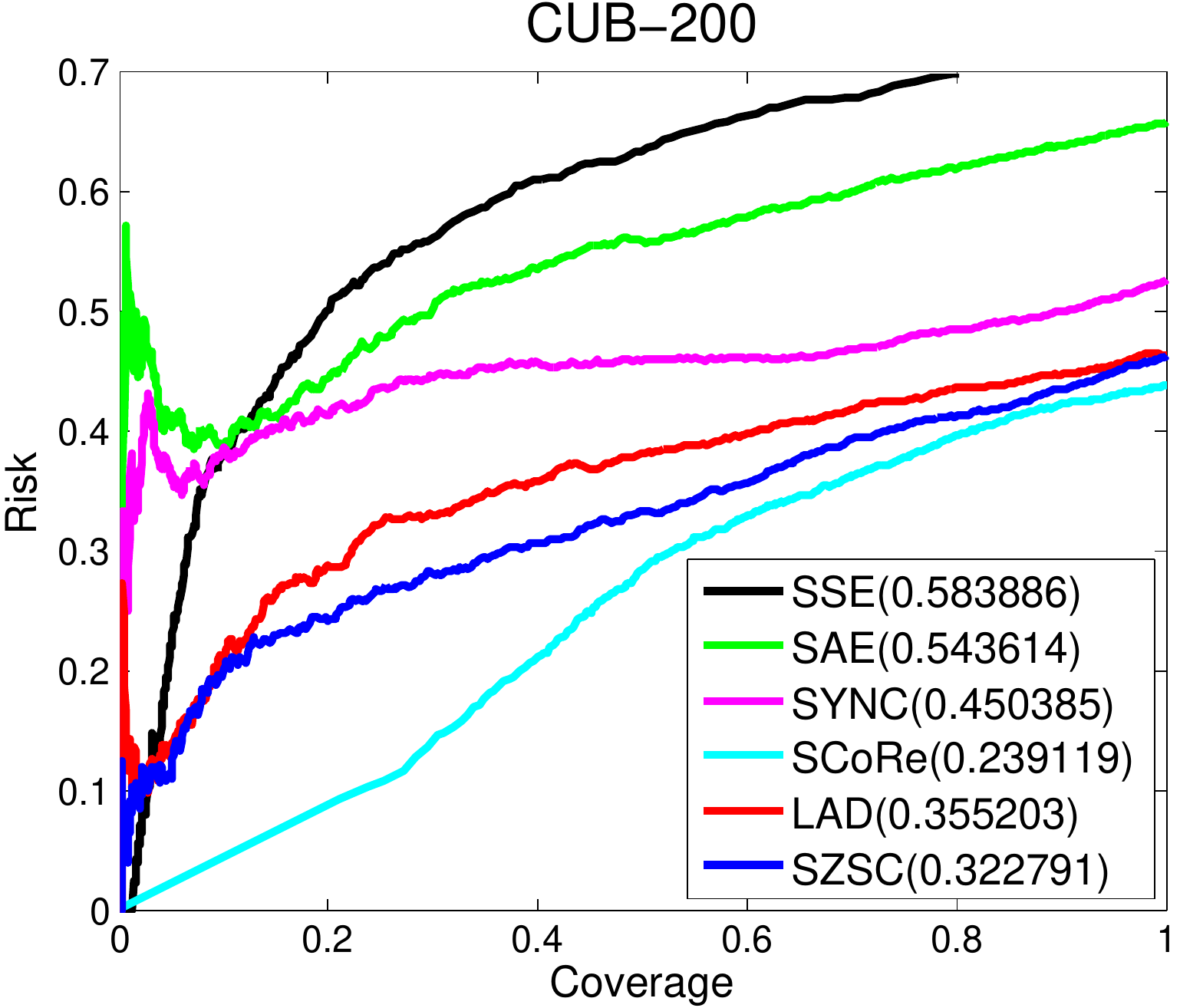}
  \caption{Risk-coverage curves of existing methods and SZSC.}
  \label{fig:competitors_review}
\end{figure}

For fair comparisons, both the proposed method and the competitors are tested with features extracted by VGG19. Experimental results are provided in Fig.~\ref{fig:competitors_review}. From the figure, we can conclude that: (1) Many existing ZSC methods exhibit poor performance in Selective ZSC settings. With lower coverage, these classifiers are expected to yield higher accuracy (i.e., lower risk). However, many methods violate that regularity in many cases, especially on aP\&Y and CUB. These experimental results give us a more comprehensive view of existing ZSC methods. (2) The proposed method outperforms most existing methods significantly on all the three benchmark datasets. One exception is that SCoRe which utilizes deep neural networks behaves better on CUB-200. However, it produces a much worse performance on aP\&Y, as there is a large imbalance among the number of images in different categories ($51\sim5071$). (3) Although bringing some improvement, the proposed method remains far behind the ideal. It indicates that there still exists large space for further study.

\subsubsection{Augmenting the Self-awareness of  Existing Methods.} 
\begin{figure}[t]
  \centering
  \includegraphics [scale=0.20]{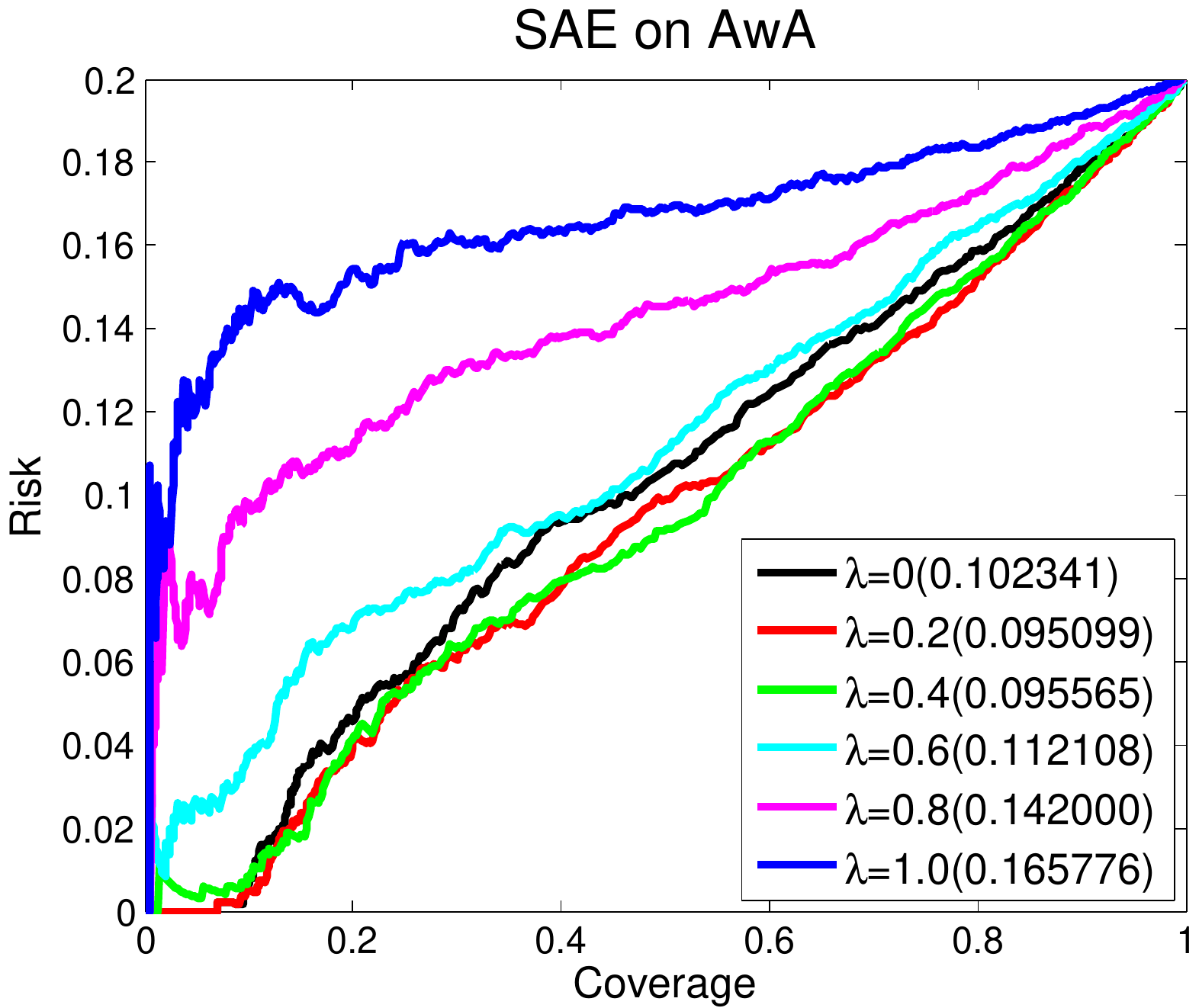}
  \includegraphics [scale=0.20]{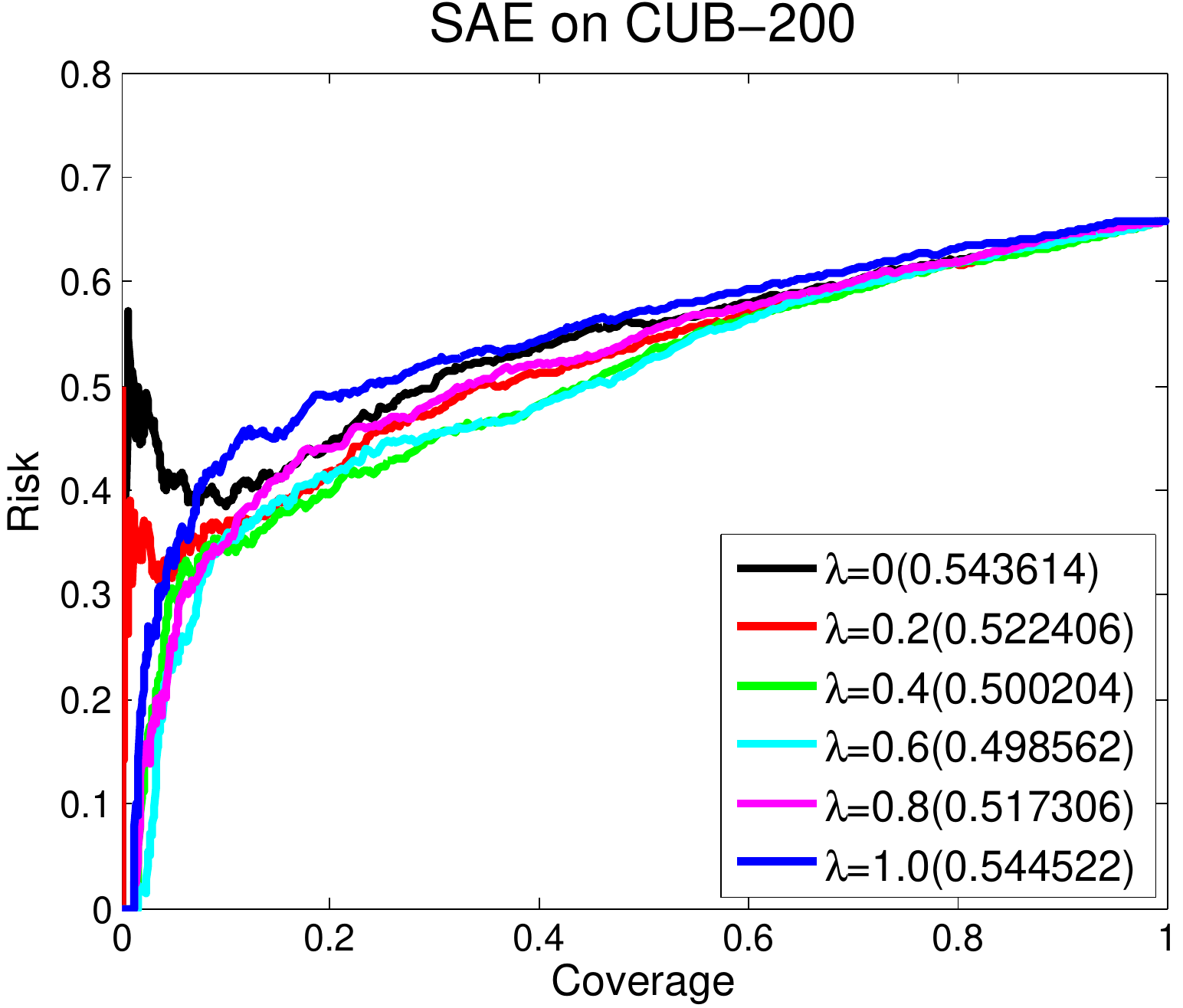}
  \includegraphics [scale=0.20]{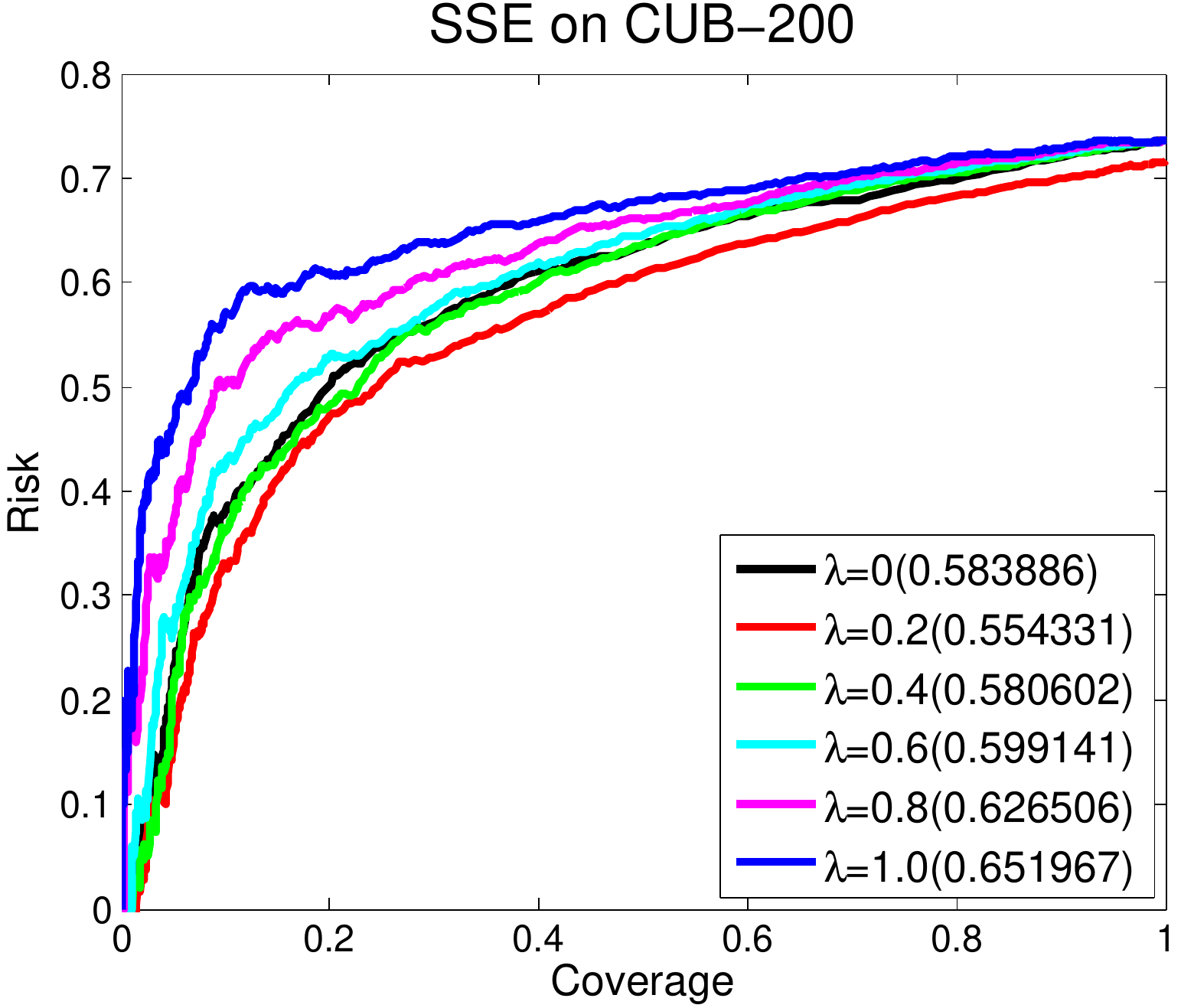}
  \caption{Combining SZSC with SAE and SSE.}
  \label{fig:com}
\end{figure}

The proposed method focuses on augmenting the defined attributes with residual properties to improve zero-shot performance in selective classification settings. It is orthogonal to how to exploit the defined attributes for ZSC. Thus the proposed method can be combined with most existing attribute-based methods to improve their performance in Selective ZSC settings. Here we propose a simple combining strategy: the confidence functions of existing ZSC methods are directly combined with the proposed confidence function defined with the aid of residual attributes. In other words, $conf_r$ is agnostic about the classification model which is used for recognition, and $conf_d$ in Eqn.~\ref{eq:confidence} is replaced with the confidence of existing ZSC methods. Experiments are conducted with SAE on AwA and CUB-200 and SSE on CUB-200. Results are shown in Fig.~\ref{fig:com}. We can see that with the simple proposed combining strategy, the performance of SAE and SSE can be further improved to some degree. These compelling results suggest that the confidence defined by the consistency between the defined and the residual attributes has some generalization ability across ZSC models. We believe learning the residual attributes adaptively with the specified ZSC model (\textit{e.g.}, SCoRe) will further improve the performance, which is left for future research.

\section{Conclusions and Future Work}
In this paper, we introduce an important yet under-studied problem: zero-shot classifiers can abstain from prediction when in doubt. We empirically demonstrate that existing zero-shot classifiers behave poorly in this new settings, and propose a novel selective classifier to make safer predictions. The proposed classifier explores and exploits the residual properties beyond the defined attributes for defining confidence functions. Experiments show that the proposed classifier achieves significantly superior performance in selective classification settings. Furthermore, it is also shown that the proposed confidence can also augment existing ZSC methods for safer classification.

There are several research lines which are worthy of further study following our work. For example, we propose to learn residual attributes to improve the performance of attribute-based classifiers. Similar ideas may also work for zero-shot classifiers built on word vectors or text descriptions. Another example is that in this paper we propose a straightforward combing strategy to improve the performance of existing methods. We believe learning the residual attributes adaptively with the ZSC model can further improve the final performance. Finally, considering the importance of the proposed selective zero-shot classification problem, we encourage researchers to pay more attention to this new challenge.\\
\\
{\footnotesize \textbf{Acknowledgements.} This work is supported by National Key Research and Development Program (2016YFB1200203), National Natural Science Foundation of China (61572428,U1509206), Fundamental Research Funds for the Central Universities (2017F\\ZA5014) and Key Research, Development Program of Zhejiang Province (2018C01004) and ARC FL-170100117, DP-180103424 of Australia.}

%\clearpage

\bibliographystyle{splncs}
\bibliography{egbib}
\end{document}